\documentclass{article}

\usepackage{arxiv}

\usepackage[utf8]{inputenc} 
\usepackage[T1]{fontenc}    
\usepackage{hyperref}       
\usepackage{url}            
\usepackage{booktabs}       
\usepackage{amsfonts}       
\usepackage{nicefrac}       
\usepackage{microtype}      
\usepackage{lipsum}		
\usepackage{graphicx}
\usepackage{natbib}
\usepackage{doi}

\usepackage{hhline}	
\usepackage{color,xcolor,ulem}
\usepackage{caption}
\usepackage{varwidth}
\usepackage{comment}
\usepackage{verbatim}
\usepackage{subfig}
\usepackage{bbm}
\usepackage{amsmath}
\usepackage{commath}
\usepackage{multirow}
\usepackage{tabularx}

\usepackage{algpseudocode,algorithm}

\title{MiSuRe is all you need to explain your image segmentation}

\author{ \href{https://orcid.org/0000-0002-5915-4528}{\includegraphics[scale=0.06]{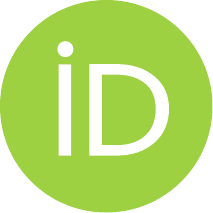}\hspace{1mm}Syed Nouman Hasany} \\
	Universit\'e de Rouen Normandie \\
	\texttt{syed-nouman.hasany@univ-rouen.fr} \\
	\And
\href{https://orcid.org/0000-0002-8656-9913}{\includegraphics[scale=0.06]{orcid.pdf}\hspace{1mm}Fabrice M\'eriaudeau} \\
Universit\'e de Bourgogne \\
\texttt{fabrice.meriaudeau@u-bourgogne.fr} \\
    \And
\href{https://orcid.org/0000-0003-0013-5370}{\includegraphics[scale=0.06]{orcid.pdf}\hspace{1mm}Caroline Petitjean} \\
Universit\'e de Rouen Normandie \\
\texttt{caroline.petitjean@univ-rouen.fr} \\
}

\renewcommand{\shorttitle}{\textit{arXiv} Template}

\hypersetup{
pdftitle={MiSuRe is all you need to explain your image segmentation},
pdfsubject={cs.AI, cs.CV},
pdfauthor={Syed Nouman Hasany, Fabrice M\'eriaudeau, Caroline Petitjean},
pdfkeywords={XAI, explainability, image segmentation, saliency, CNN, transformer, sufficient region},
}

\begin{document}
\maketitle

\begin{abstract}
The last decade of computer vision has been dominated by Deep Learning architectures, thanks to their unparalleled success. Their performance, however, often comes at the cost of explainability owing to their highly non-linear nature. Consequently, a parallel field of eXplainable Artificial Intelligence (XAI) has developed with the aim of generating insights regarding the decision making process of deep learning models. An important problem in XAI is that of the generation of saliency maps. These are regions in an input image which contributed most towards the model's final decision. Most work in this regard, however, has been focused on image classification, and image segmentation - despite being a ubiquitous task - has not received the same attention. In the present work, we propose MiSuRe (\textbf{Mi}nimally \textbf{Su}fficient \textbf{Re}gion) as an algorithm to generate saliency maps for image segmentation. The goal of the saliency maps generated by MiSuRe is to get rid of irrelevant regions, and only highlight those regions in the input image which are crucial to the image segmentation decision. We perform our analysis on 3 datasets: Triangle (artificially constructed), COCO-2017 (natural images), and the Synapse multi-organ (medical images). Additionally, we identify a potential usecase of these post-hoc saliency maps in order to perform post-hoc reliability of the segmentation model.
\end{abstract}

\keywords{XAI \and explainability \and image segmentation \and saliency \and CNN \and transformer \and sufficient region}

\section{Introduction}
\label{sec:intro}

Following the success of AlexNet \citep{alexnet} in 2012 in the ImageNet competition, Deep Learning based algorithms - particularly CNNs and, more recently, transformers - swiftly dethroned the classical approaches in a variety of standard computer vision tasks. Multiple reasons have contributed to this success such as the availability of sufficient amounts of data as well as compute resources. Another such reason is the multi-layer, highly non-linear nature of these algorithms which allows them to model complex relationships given enough data. While this non-linearity affords high predictive performance, it also leads to the model not being inherently interpretable. Such models are often labelled as black-box models as opposed to the comparatively transparent white-box models (for example: decision trees) where the model is inherently interpretable allowing us useful insights as to why the model arrived at a particular decision. White-box models, however, are often too simplistic for computer vision problems, and fail to achieve reasonable performance critera.

Naturally, a parallel field has since developed within Deep Learning which focuses on explaining model decisions. For the end-user, these explanations help improve their confidence in the model, whereas for a developer these explanations can help in a variety of ways such as identifying bias in the model. When it comes to explaining individual model decisions (post-hoc local interpretability), two broad streams can be identified in deep learning: (i) \textbf{gradient} based, and (ii) \textbf{perturbation} based. Gradient based techniques \citep{simonyan,SG,IG,Selvaraju_2019} utilize the availability of the gradient associated with the model's output with respect to the input image or to any of the intermediate activation maps. Perturbation based approaches \citep{fergus,lime,shap,rise}, on the other hand, rely on modifying the input image, and the corresponding change in the model's output as a result of this modification. For a given model decision, both these streams output a saliency map indicating the region(s) in the input image which was supposedly the model's focus.


Image classification has received the most attention when it comes to explainability approaches in deep learning based computer vision. Explainability approaches span a diverse set of tasks such as saliency maps, concept identification, counterfactual explanations, etc. Saliency maps are maps in which regions in an input image are highlighted in proportion to their contribution towards the final decision. Dense prediction tasks such as image segmentation have received far less spotlight in this regard. Some of the saliency generation approaches which have been developed for image segmentation are extensions of approaches proposed in the context of image classification where the only modification is made in terms of dealing with the model's output - single valued prediction as opposed to a dense prediction - while the rest of the approach remains unchanged. A possible factor which has led to image segmentation receiving less attention is the relative lack of need for a saliency map when it comes to segmentation as opposed to classification. As the output of an image segmentation is a delineation of objects present in the input image, one can be led to the conclusion that the saliency in this case would simply be the individual objects themselves, hence rendering the need for an external process to generate explanations redundant \citep{context-bias}.

Most of the methods proposed in the context of saliency generation in image segmentation are \textbf{gradient} based approaches. Specifically, they are extensions of Grad-CAM \citep{gradcam} and its derivatives. A much smaller group belongs to the \textbf{perturbation} based approaches such as extension of RISE to image segmentation \citep{dardouillet:hal-03719597}. Certain flaws can be identified in both these streams. When it comes to Grad-CAM in image classification, it is usually applied to the final feature extraction convolutional layer. This layer contains an effective summary of the image, and is solely responsible for the final model output. When it comes to extending Grad-CAM to image segmentation models such as U-Net, the choice of layer becomes non-trivial. Whereas the bottleneck can be thought of as the layer containing a summary of the image, it is hardly the only layer responsible for the final output decision. On the other hand, the layer responsible for the final output decision (just prior to the segmentation head) does not contain a summary of the original image. Even though most practitioners \citep{Vinogradova_2020} choose to generate the saliency map from the bottleneck layer, this choice is hard to justify. Additionally, CAM-based algorithms \citep{cam} were designed for convolutional architectures and are not guaranteed to work for other architectures such as those involving transformers \citep{transunet, unetr}. Neither of these problems is encountered with perturbation based approaches as they are model agnostic. Perturbation based approaches such as RISE, however, tend to return a coarse saliency map which does not allow for the identification of precise locations of importance. The present work aims to solve both of these challenges.

In the present work, we propose a two stage method for generating saliency maps in image segmentation. In the first stage, motivated by the inductive bias inherent in the segmentation process, a mask is initialized focused on the object under consideration. This mask is gradually dilated until the segmentation model can successfully segment the object. The saliency map generated from this first stage represents a sufficient region which the model requires for a successful segmentation. In the second stage, taking inspiration from the work of Fong and Vedaldi \citep{min_pert} in image classification, we optimize over the mask directly to prune this sufficient region in order to approximate a minimally sufficient region - where an ideal minimally sufficient region would only retain those portions of the input image without which the model will not be able to arrive at the correct segmentation. This approach ends up providing two saliency maps, where the sufficient region represents a coarser version of the explanation, and the minimally sufficient region represents a finer version. We also utilize our method to explore image segmentation in general in order to extract further insights with regards to the overall segmentation process. Finally, we explore the relationship between the generated saliency maps and the ground truth mask. This opens up the possibility for the post-hoc determination of reliable predictions from unreliable ones.

Our contributions can be listed as follows:
\begin{itemize}
    \item We propose a model-agnostic two stage method for generating saliency maps in image segmentation, and showcase its application on three diverse datasets, one artificial (Triangle Dataset), one medical (Synapse multi-organ CT), and one natural (COCO-2017), both with convolutional and transformer based architectures ;
    \item We utilize our model to extract further insights with respect to the image segmentation process, regarding the size of the object to be segmented and the nature of the image (natural images vs. medical images);
    \item We identify a potential application of our saliency generation process as a proxy for post-hoc model reliability.
\end{itemize}

\section{Related Work}
\label{sec:rw}

\subsection{Explainability in  Image Classification}

One of the earliest works in interpretability was of Zeiler and Fergus in 2013 \citep{fergus} in which they introduced a perturbation based technique, occlusion, as a means of generating a saliency map. Following that, a gradient based technique was proposed by Simonyan et al. \citep{simonyan} in which a saliency map was defined as the gradient of the output score with respect to the input image. This gradient based technique was further refined in subsequent works in which the primary focus was towards cleaning the otherwise noisy gradient information. Examples of such techniques include SmoothGrad \citep{SG} and IntegratedGradients \citep{IG}. Instead of propagating the gradient all the way back to the input image, some techniques chose to rely on the intermediate activation space in order to generate a saliency map. Grad-CAM \citep{gradcam} is one such approach in which a linear combination of activation maps from an intermediate network layer is considered as an explanation. The linear coefficients of this linear combination are obtained from the gradient of the output score with respect to the activation maps under consideration. Grad-CAM inspired many derivatives such as Grad-CAM++ \citep{gradcam++} and LayerCAM \citep{layercam}. Certain derivatives, however, identified the unreliability of using gradient information to compute the linear combination coefficients, and suggested alternate methods such as Score-CAM \citep{scorecam} and Ablation-CAM \citep{ablationcam}. Where the original Grad-CAM utilized a single backward pass, these derivatives, however, required multiple passes, and coupled with the fact that they were not operating directly on the input image made them considerably less attractive compared to the original technique.

A number of perturbation based techniques have also been proposed which generally operate directly on the input image by modifying it, and observing the changes of the modification on the model's output. Other than occlusion \citep{fergus}, LIME \citep{lime}, SHAP \citep{shap}, and RISE \citep{rise} are popular techniques which generate multiple modified instances of the original image, and model the network's behavior on these modifications using an inherently interpretable model. The generated modifications, however, are mostly random and do not take the network's behavior on the input image under consideration. An alternate approach was propsed by Fong and Vedaldi \citep{min_pert} which iteratively perturbs the input image guided by the gradient information - meaningful perturbations. The goal is to delete regions in an image which are maximally informative, and the removal of which would lead to the model changing its prediction. Alternatively, the goal can also be to end up with the minimally sufficient region of the input image which is necessary for the model to preserve the correct prediction.

\subsection{Explainability in Image Segmentation}

An early work which extended Grad-CAM to image segmentation was Seg-Grad-CAM by Vinogradova et al \citep{Vinogradova_2020}. Unlike image classification where there is a single output value, segmentation is a dense prediction task. In order to take this into account, a sum of scores from the region under consideration was taken as the output, the gradient of which was utilized to compute the linear coefficients. As the choice of network layer to generate a saliency map is not obvious in image segmentation \footnote{The original paper utilized the bottleneck layer of a U-Net model}, Mullan and Sonka proposed combining activation maps from each stage of the decoder instead of relying on a single layer \citep{visual_attribution_seg}. Additionally, it was identified that Grad-CAM's utilization of global average pooling to compute the linear coefficients might not work well when it comes to explaining a spatially local region in a segmented image \citep{segnbdt, segxrescam}. 

In terms of perturbation based techniques, occlusion \citep{occ_seg}, SHAP \citep{dardouillet:hal-03719597}, and RISE \citep{dardouillet:hal-03719597} have been extended to image segmentation whereby, for the latter two, a linear combination of the masks involved in generating the multiple modifications of the original input image serve as the explanation for the model's output. Additionally, the meaningful perturbations approach was extended to image segmentation in which the input image is iteratively perturbed with the help of gradient information \citep{grid}. In this approach, however, perturbations are not applied to the segmented object to be explained leading to a contextual explanation.

While these methods are post-hoc locally interpretable methods, some work has also been done on global explainability in image segmentation. Janik proposed dimensionality reduction, and subsequent visualization of the bottleneck latent space as well as the identification of favorable and unforavorable regions for the model in this latent space \citep{janik}. Koker et al. proposed training a deep neural network for the sole purpose of predicting the saliency map guided by the original segmentation model which is frozen \citep{u-noise}.

Our work takes inspiration from both Fong and Vedaldi's \citep{min_pert} approach of meaningful perturbation as well as Hoyer et al.'s \citep{grid} extension of their approach to generate contextual explanations in image segmentation. Our departure from Hoyer et al.'s is primarily in three aspects. First, we explore the generation of a saliency map instead of a contextual explanation, and in doing so we allow perturbation to impact the object under consideration as well. This allows us to investigate whether the entirety of the object is, in fact, necessary for its successful segmentation. Second, we utilize the inductive bias inherent in the image segmentation process in order to initialize and refine our mask (using dilations) to arrive at a sufficient region before initializing the optimization based perturbation process. Third, our optimization goal is to maximize the Dice score between the model's prediction on the perturbed image and its prediction on the original image as opposed to minimizing the $l_{1}$ norm between the two predictions.   

\subsection{Evaluation}

A popular technique to evaluate saliency maps in image classification works by gradually adding portions of the image, starting with the most salient and ending with the least salient, such that by the end of the sweep the entire image content has been added. During this sweep the progression of the model's output is tracked, and ideally, the model's output should increase significantly as soon as the most salient regions of the image are added. This can be extended to image segmentation, and instead of the model's output, one can utilize the Dice score between the prediction of the modified image and the prediction on the original image. However, Mullan and Sonka \citep{visual_attribution_seg} recently proposed another technique specifically for image segmentation in which two scores are calculated, one corresponding to the segmentation performance on the input image masked by the saliency map - prediction preserved, and the second corresponding to the percentage of image pixels in the saliency map - image preserved. A saliency map preserving a smaller percentage of the image is better than one preserving a larger percentage given a similar segmentation performance. 

\section{Method}
\label{sec:metj}

\subsection{Generation of Sufficient Region ($X_{SR}$) and Minimally Sufficient Region ($X_{MSR}$)}

Let us consider a segmentation network $f_{\theta}$ parameterized by $\theta$. Let us consider an image $X_{0}$ defined over a domain $\Omega \subset \mathbb{R}^2$ ($|\Omega| = N$ pixels) with values in $\mathbb{R}^{C}$ where $C$ is the number of channels, and its associated segmentation map $Y_{0}$ defined over domain $\Omega$ with values in ${0, 1, ..., L-1}$ where $L$ is the number of classes.

Let $l$ be the class for which a saliency map has to be generated. Let $M$ be a mask of spatial dimensions $H \times W$ initialized as all ones. Given the inductive bias inherent in the image segmentation process, we initialize the mask with the assumption that the most salient region would  encompass the segmented object itself. The mask is first resized to match the spatial dimensions of the image. Following that, we switch off our mask\footnote{switch off implies zeroing the values} on all spatial locations other than those corresponding to where the segmentation model predicted the class $l$ for $X_{0}$. This leads to an initial mask $M_{0}$ such that : $M_{0} = \mathbbm{1}(Y_{0})$ where $\mathbbm{1}()$ is the indicator function, equal to 1 when  $Y_{0}=l$ and 0 otherwise. Then the initialized mask $M_{0}$ is elementwise multiplied with the input image $X_{0}$ i.e.\ $X_{m} = X_{0} \odot M_{0}$, where $\odot$ refers to the Hadamard product. We use this $X_{m}$ as the input to the segmentation model, and compute a Dice score between the model's prediction on $X_{m}$ denoted  $f_{\theta}(X_{m})$  and the model's prediction on $X_{0}$ for the category $l$.

 We define a threshold $\tau$ ($0.9$ in our case), and if the Dice score is above that, we consider $M_{0}$ to be the mask corresponding to a sufficient region. In case the Dice score (DSC) is less than the threshold, we dilate our mask, and repeat the process until the Dice score is above the threshold. The mask $M_{SR}$ obtained from this process is elementwise multiplied with the input image in order to generate the region that we refer to as the "sufficient region" (SR) i.e.\ $X_{SR} = X_{0} \cdot M_{SR}$.\footnote{in a case where $M_{0}$ requires zero dilations, $M_{0}$ is identical to $M_{SR}$}. This method is summarized in algorithm \ref{msr_algo}.

\begin{algorithm}
 \caption{Finding $M_{SR}$}

\begin{algorithmic}[1]
\Function{Finding $M_{SR}$}{} :

    $M_{SR} = M_{0}$

    \While {$DSC(f_{\theta}(X_0),f_{\theta}(X_m)) \leq \tau$}
        \State $M_{0} \longleftarrow dilate (M_{0})$
        \State $X_m \longleftarrow X_{0} \odot M_{0}$
    \EndWhile
    \State \Return $X_{SR} = X_{m}$, $M_{SR} = M_{0}$
\EndFunction
\end{algorithmic}
\label{msr_algo}
\end{algorithm}

In the second step, our goal is to refine $X_{SR}$ in order to find the minimally sufficient region $X_{MSR}$. In order to achieve this, we define the following optimization objective:

\begin{equation}    
\begin{split}
    \label{eq-msr}
    M_{MSR} & = \underset{M_{SR}}{\arg \min} \ \frac{\lambda}{|\Omega|} \sum_{u \in \Omega} \norm{M_{SR}(u)}_{1} + 
    \gamma \sum_{u \in \Omega} \norm{\nabla M_{SR}(u)}_{\beta}^{\beta}\\
    & + \left \{1 - \sum_{i \in \{0, l\}}{\alpha_{i}}\frac{\sum_j^N 2f_{\theta}(X_{SR} \odot M_{SR})_{i}^jf_{\theta}(X_{0})_{i}^j + \epsilon}{\sum_j^Nf_{\theta}(X_{SR} \odot M_{SR})_{i}^j + \sum_j^Nf_{\theta}(X_{0})_{i}^j + \epsilon} \right \}
\end{split}
\end{equation}

The first term with $\lambda$ as its hyperparameter is the absolute average of the $M_{SR}$, the goal of the optimization is to decrease this as much as possible with the aim of removing unnecessary regions from the mask. The second term with $\gamma$ as its hyperparameter is the total variation (TV) regularization which contributes towards the mask being smooth. Following that we have the Dice loss. The goal is to minimize the Dice loss in catergory $l$ between the prediction on the original image and the prediction on the perturbed image where the perturbation is defined as $\phi_{SR} = X_{SR} \odot M_{SR}$.\footnote{It is indeed possible to replace Dice with other losses such as the cross entropy loss} The preservation terms for the foreground and background Dice can potentially have different coefficients with $\alpha_{l} \geq \alpha_{0}$ in order to favor the preservation of the foreground. $j$ is the index over the total number of pixels ($N$) in a channel. The output from this step (eq. \ref{eq-msr}) $M_{MSR}$ is utilized to generate our minimally sufficient region $X_{MSR}$:

\begin{equation}
    X_{MSR} = X_{SR} \odot M_{MSR}
\end{equation}

Even though $X_{SR}$ is generated to provide a better initialization for the optimization step, it can, in itself, be seen as an explanation alongside $X_{MSR}$ with the former providing a coarser explanation, and the latter providing a much finer one.

\subsection{Extracting Global Insights on the Segmentation Process}

By design, the saliency maps we generate belong to the genre of local explainability techniques in which a method is applied to explain decisions for individual data instances. Features from individual saliency maps can, however, be combined allowing for further analysis. We plot (i) the number of dilations required for $X_{SR}$ vs. the prediction size as well as the (ii) the evaluation metric vs. the prediction size. These two plots allow us to observe a general trend in the overall segmentation process.

\subsection{Post-Hoc Assessment of the Segmentation Model Reliability}

The purpose of saliency maps is usually to either increase the confidence of the end user in the model's predictions or to help the developer debug the model in order to improve its performance. In the present work, we explore the relationship of saliency maps to the post-hoc reliability of the segmentation model. Given the assumption that saliency maps for incorrect model predictions are potentially different from those of saliency maps for correct model predictions, we experiment with training a discriminator which can allow us to identify whether our model is correct in its prediction or whether it has faltered. The features utilized to train this discriminator model are extracted from the saliency maps themselves. The utility for such a discriminator is evident as it can allow us to automatically judge the model's predictions for cases where the ground truth labels are not available.

\subsection{Evaluation}

Given the inherently subjective nature of explainability, there is a lack of consensus when it comes to metrics involving saliency maps. Occasionally, works limit themselves to a qualitative analysis based on the visual results of the saliency maps. In the present work we present both a qualitative as well as a quantitative analysis based on metrics which shall briefly be discussed below.

For a quantitative metric we take inspiration from Mullan and Sonka \citep{visual_attribution_seg}. They proposed two metrics which jointly take both the segmentation performance as well as the image preservation into account when evaluating saliency maps. For the first of these metrics, a Dice score is calculated between the model's prediction on $X_{MSR}$, and the model's prediction on the original image in order to determine how well the region relevant to the prediction was captured by the saliency map. We refer to this as \textbf{Dice explained}. For the second metric, the percentage of non-zero pixels in the saliency map is computed in order to quantify the size of the image which was preserved. We modify this latter percentage, and instead calculate the ratio of the sum of non-zero pixels in the saliency map to the number of pixels in the segmented object. We refer to this as the \textbf{perturbation ratio}. A good saliency generation method is going to have a high Dice explained (as close to $1$ as possible) and a low perturbation ratio.

\section{Experiments}

\subsection{Datasets and Models}

We conduct our experiments on three diverse datasets, one of them being artificially generated, one being medical in nature, and the last consisting of natural images. The artificially generated dataset is inspired from Riva et al. \citep{florian}. This dataset, called Triangle, is generated by placing objects from the Fashion-MNIST on a blank image. Three objects are placed in a triangular fashion such that the objects' centers form the vertices of a triangle. The other objects are randomly placed. The goal of a model is to only segment those objects which are part of the triangle, and consider the randomly placed objects as being part of the background (see Fig. \ref{florian} for examples). We generate 2000 samples, 1400 of which are for training, and the remaining 600 for validation. We trained a U-Net (VGG-16 backbone) on this dataset for our saliency experimentation. 

The second dataset is the Synapse multi-organ CT dataset \citep{landman2015miccai} consisting of 30 abdominal CT scans, 18 of which are used for training and 12 for validation. In total we have 2211 2D slices for training, and 1568 slices for validation. The labels are: Aorta, Gallbladder, Left
Kidney, Right Kidney, Liver, Spleen, Pancreas, and Stomach. We trained a U-Net (ResNet-34 backbone) \citep{unet} as well as a TransUNet \citep{transunet} on this dataset for our saliency experimentation.

The final dataset is the COCO-2017 dataset \citep{coco} from which we select the following labels: Bus, Car, Cat, Cow, Dog, Bike, Person, and Train. For our saliency experimentation we utilized a pre-trained DeepLabV3 (ResNet-50 backbone) \citep{deeplab} model.

\subsection{Experimental Configuration}

For the saliency experiments, we use a single channel mask with a size of $224 \times 224$ (for mask initialization, images are accordingly resized to these dimensions). The AdamW optimizer is utilized with a learning rate of $0.1$ and the coefficient for the absolute average loss $\lambda$ (eq. \ref{eq-msr}) is 0.01. Following Hoyer et al. \citep{grid}, after every iteration of the optimization process, any value of the mask less than 0.2 is clamped to 0, and any value above $1$ is clipped to $1$. We perform the optimization for 100 iterations. In order to generate the $M_{SR}$, we use a circular dilation kernel of size $7 \times 7$. We use a coefficient of $\alpha_{0} = 1$ for the background Dice and a coefficient of $\alpha_{l} = 2$ for the foreground Dice. We also present results for alternate parameter combinations in order to contextualize our parameter choices.

\subsection{Evaluation}

For a qualitative analysis, sample saliency maps are shown. For a quantitative analysis, our primary evaluation metrics are the \textbf{Dice explained} and the \textbf{perturbation ratio}. We report the mean performance of our saliency method for both these metrics on a subset of our utilized datasets.

\subsection{Comparison to Baselines}

We compare our saliency method against the most popular saliency generation method in image segmentation, Seg-Grad-CAM \citep{Vinogradova_2020}, a gradient based method. We also compare it to a perturbation based method, RISE \citep{dardouillet:hal-03719597}, as it is also a model-agnostic saliency generation method. RISE works on a perturbation based scheme whereby randomly generated masks are applied to the input image; and for each of those masked images, a Dice score is calculated between the prediction on the masked image and the prediction on the original image. A linear combination of the generated masks serves as our explanation where the coefficients are the aforementioned Dice scores. For our experiments we use Seg-Grad-CAM on the bottleneck layer, and RISE with $2000$ masks. 

We compare both qualitatively (based on the visual saliency map) as well as quantitatively using the Dice explained and perturbation ratio metrics. Saliency maps from both RISE and Seg-Grad-CAM are first thresholded followed by an elementwise-multiplication of the binarized saliency map with the original image. For Seg-Grad-CAM, thresholds of 0.05 and 0.1 are experimented with, whereas for RISE, thresholds of 0.2 and 0.4 are utilized.

\section{Results and Discussion}

\subsection{Sample Saliency Maps}\label{ssm}

\begin{figure*}[t]
    {\includegraphics[width=1\textwidth]{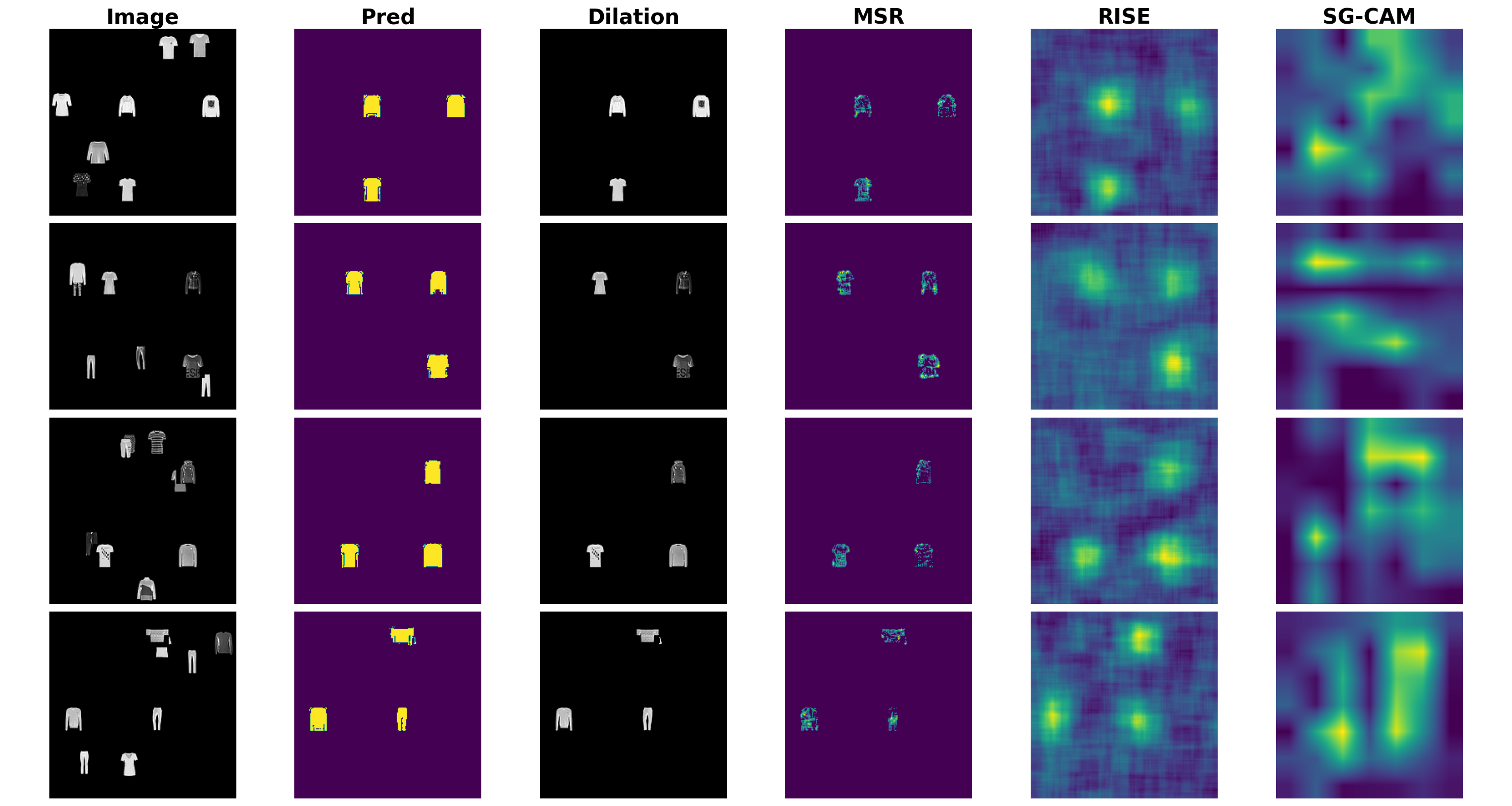}} \\
    \caption{Sample results (in row) from the Triangle dataset. The column 'Dilation' refers to $X_{SR}$ whereas column 'MSR' (saliency map) refers to $M_{MSR}$. SG-CAM: Seg-Grad-CAM. Results are best viewed zoomed-in.}
    \label{florian}
\end{figure*}

Figure \ref{florian} shows saliency maps generated for a few samples of the Triangle dataset. In all cases, it is apparent that the saliency map generated by MSR and RISE broadly agree in their localization. Seg-Grad-CAM, on the other hand, has generated significantly divergent saliency maps. Between MSR and RISE, the main difference lies in the map's fineness. MSR's maps are considerably finer in which the most prominent regions appear to be the boundaries of the objects. For RISE, however, saliency maps are fairly coarse, and despite them being most active for the three objects, they fail to disclose any further information with regards to delineating the exact region important for the segmentation model.

\begin{figure*}[t]
    {\includegraphics[width=1\textwidth]{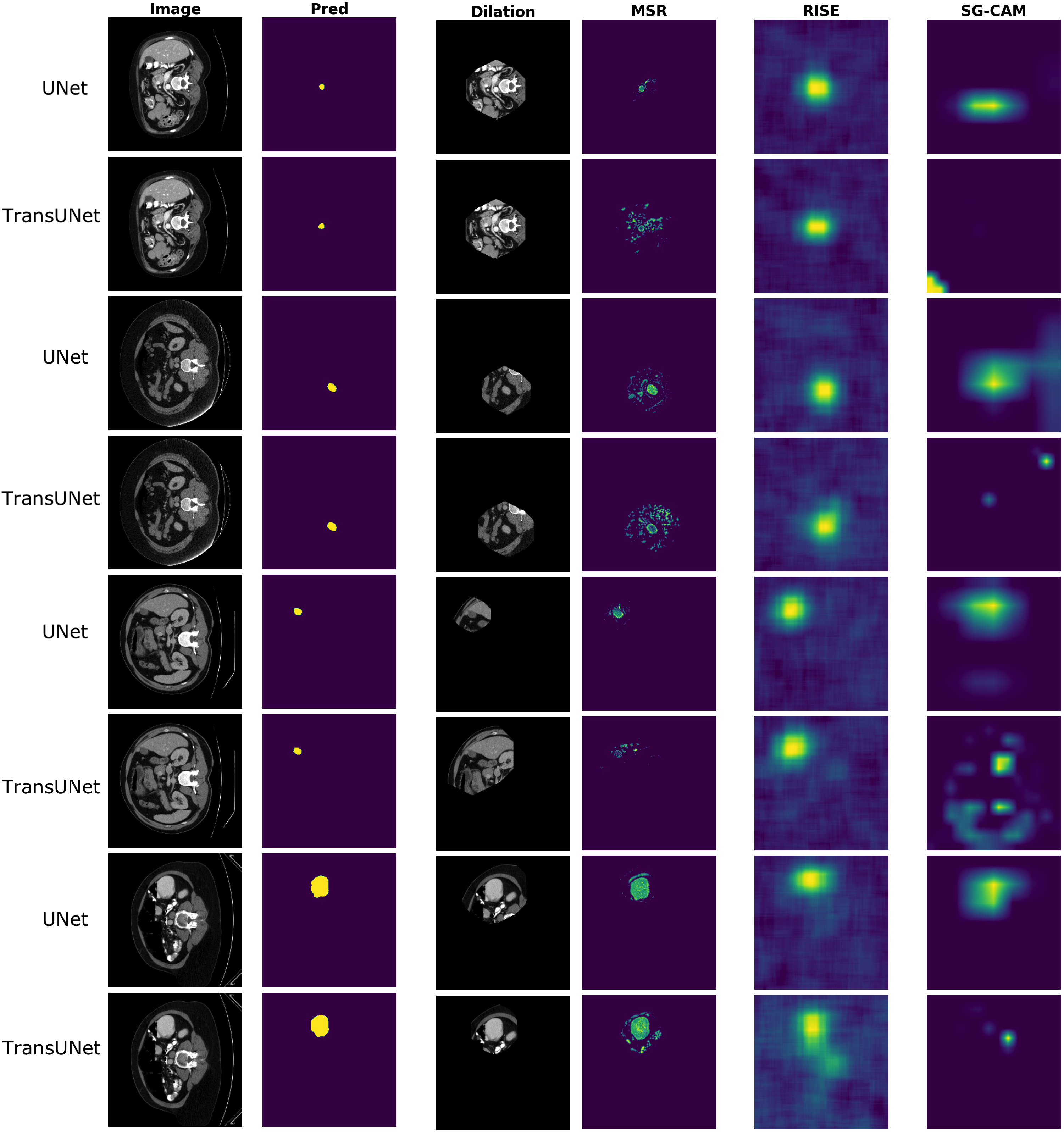}} \\
    \caption{Sample results from the Synapse multi-organ CT dataset, from U-Net and TransUNet. Each pair of rows (top-to-bottom) represents a class to be explained: Aorta, Left Kidney, Gall Bladder and Liver. Dilation refers to $X_{SR}$ whereas MSR (saliency map) refers to $M_{MSR}$. Results are best viewed zoomed-in.}
    \label{synapse}
\end{figure*}

Figure \ref{synapse} shows saliency maps generated for a few samples on the Synapse dataset. Once again, we see agreement in localization between MSR and RISE, independently from the model (U-Net vs. TransUNet), with the primary difference in saliency maps being that of fine vs. coarse. In the case of MSR, a comparison of the \textbf{Dilation} column to the \textbf{MSR} column also allows us to see the importance of optimization following the dilation step as it leads to non-essential information being pruned away.
For Seg-Grad-CAM, there is a big discrepancy, depending on the model: where saliency maps obtained on UNet may show some localization correspondence with RISE, those obtained on TransUNet are highly questionable, and highlight the inadequacy of Seg-Grad-CAM for non-convolutional architectures.

\begin{figure*}[t]
    {\includegraphics[width=1\textwidth]{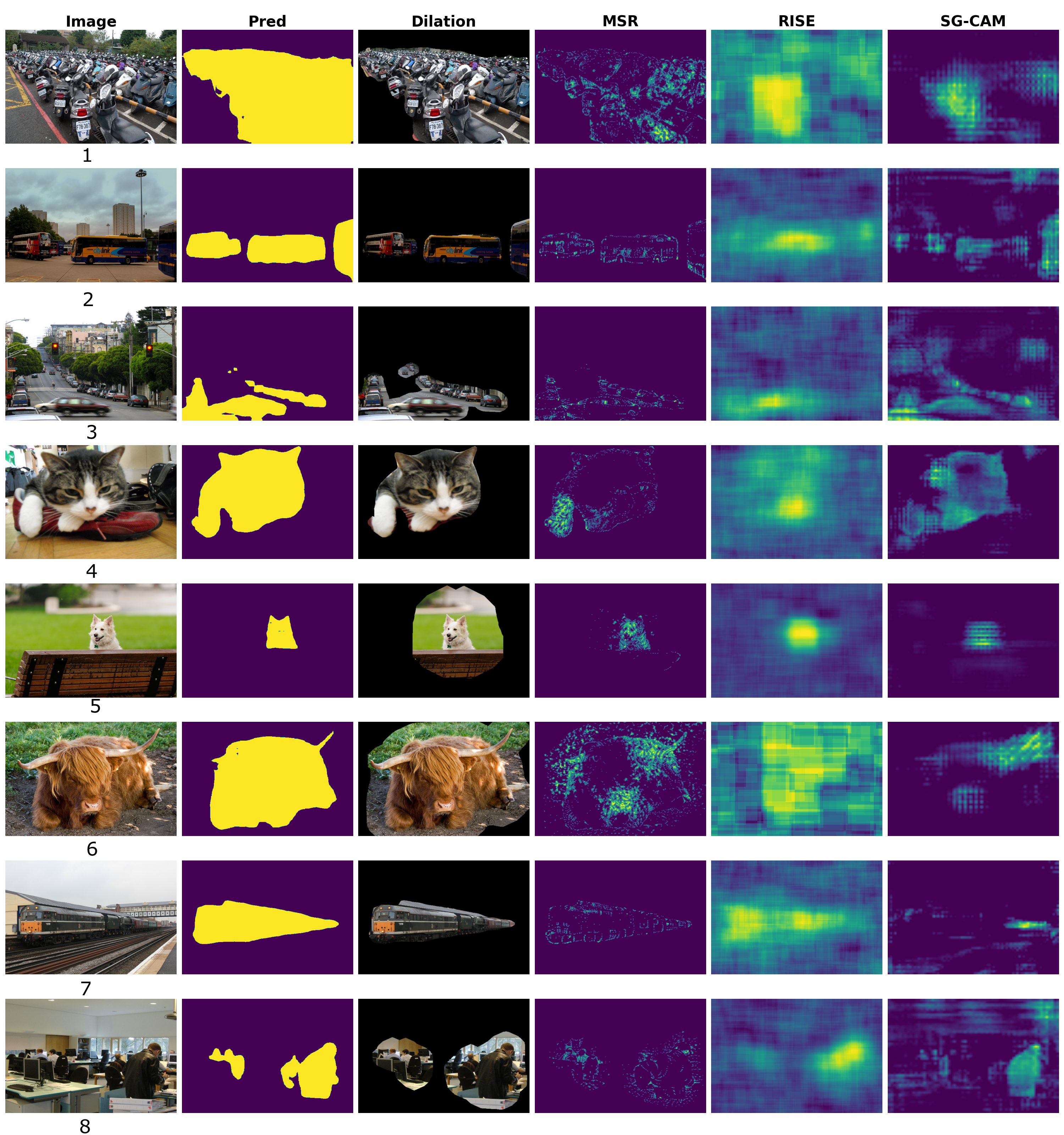}} \\
    \caption{Sample results from the COCO-2017 dataset. Row number is below the image. Each row (top-to-bottom) represents a class to be explained: Bike, Bus, Car, Cat, Dog, Cow, Train and Person. Dilation refers to $X_{SR}$ whereas MSR (saliency map) refers to $M_{MSR}$. Results are best viewed zoomed-in.}
    \label{coco}
\end{figure*}

Figure \ref{coco} shows saliency maps generated for a few samples on the COCO-2017 dataset. In this case, all three methods seem to broadly agree on localization. From a spectrum of coarse to fine, RISE returns the coarsest saliency maps, and MSR returns the finest, with Seg-Grad-CAM being in between. Some interesting samples are from rows five, six, and seven where results from MSR and Seg-Grad-CAM not only agree on the overall localization, but also on the general shape. Row seven is particularly interesting as both MSR and Seg-Grad-CAM only highlight the boundaries of the object (train) while ignoring the remaining portion of the object almost entirely. 

Focusing on the MSR, we observe that for both the Triangle dataset as well as the COCO-2017 dataset, the saliency maps tend to indicate that the segmentation models are highly influenced by the objects' boundaries whereas for the Synapse dataset, the object itself appears to be more important for the segmentation model. For our particular case it appears that segmentation models rely on different features when it comes to segmenting natural images as compared to medical images. In the case of natural images, the contours appear as the most important feature whereas for medical images, it is the entire object itself instead of its mere boundary. While this observation requires further experiments in order to arrive at a general conclusion, medical and natural images being of a different nature in terms of their intrinsic dimensions \citep{konz2022intrinsic} might explain this difference in the segmentation models' behavior.


\clearpage

\subsection{Dice Explained and Perturbation Ratio}

Table \ref{tab:comparisons} records the results on a subset of the Triangle, Synapse, and COCO-2017 datasets. In terms of Dice explained, RISE (with a binary threshold of 0.2) consistently outperforms the other methods with MSR following it. Whereas when it comes to perturbation ratio, MSR is the clear winner. It can also be observed that the perturbation ratios of MSR are less than the others by an order of magnitude. 

Interestingly, for both the Triangle as well as the COCO-2017 datasets, the perturbation ratios obtained from MSR are less than $1$. This means that the necessary region required by the segmentation model in order to segment an object contained \textit{fewer} pixels than the object itself - two-thirds in the case of Triangle and one-fifth in the case of COCO-2017. The perturbation ratio obtained from MSR for Synapse, on the other hand, is more than $1$. For the Triangle dataset, it can be surmised that the objects and the backgrounds are simple enough such that the model can segment the entire object merely by identifying the outline. For Synapse, given that the objects to be segmented are often similar in texture to their surroundings, the required context is naturally greater. These hypothesis, however, require further investigation.

Compared to RISE and MSR, Seg-Grad-CAM does not obtain good results. The only dataset for which Seg-Grad-CAM performs comparatively well is the COCO-2017 dataset. This corroborates well with the visual results presented in Section \ref{ssm} (Figure \ref{coco}). We also observe that in Synapse's case, Seg-Grad-CAM's application on TransUNet is worse than Seg-Grad-CAM's application on U-Net. This reinforces the notion that caution should be exercise while importing Grad-CAM based algorithms to transformer based architectures.

In terms of computational performance, Seg-Grad-CAM is the quickest taking less than a second whereas RISE is the slowest taking around a minute. MSR takes around $3$ to $4$ seconds to process a single image. However, given the other two performance metrics, it is clear that Seg-Grad-CAM can hardly be the ideal contender for generating saliency maps in image segmentation. If all metrics are taken into account, MSR emerges as the most balanced of the three algorithms.

Finally, it is worth noting that the thresholds have a considerable impact on both RISE and Seg-Grad-CAM. With a smaller threshold, more pixels from the input image are included. Naturally, this leads to a better Dice explained at the expense of a worse perturbation ratio. MSR requires no threshold as its values are determined directly by the optimizer. 

\begin{table*}[]
\centering
\resizebox{1.0\columnwidth}{!}{%
\begin{tabular}{|c|c|c|c|c|c|c|}
\hline
\textbf{Dataset} & \textbf{Model} & \textbf{\begin{tabular}[c]{@{}c@{}}XAI\\ Method\end{tabular}} & \textbf{\begin{tabular}[c]{@{}c@{}}Binary \\ Threshold\end{tabular}} & \textbf{\begin{tabular}[c]{@{}c@{}}Dice \\ Explained\end{tabular}} & \textbf{\begin{tabular}[c]{@{}c@{}}Perturbation\\ Ratio\end{tabular}} & \textbf{Time (s)} \\ \hline
Triangle & U-Net & Seg-Grad-CAM & 0.05 & 0.665 & 27.38 & 0.067 \\ \hline
Triangle & U-Net & Seg-Grad-CAM & 0.1 & 0.553 & 16.86 & 0.067 \\ \hline
Triangle & U-Net & RISE & 0.2 & \textbf{0.995} & 26.07 & 52 \\ \hline
Triangle & U-Net & RISE & 0.4 & 0.984 & 5.83 & 52 \\ \hline
Triangle & U-Net & MSR & - & 0.965 & \textbf{0.63} & 3.38 \\ \hline
\hline
Synapse & U-Net & Seg-Grad-CAM & 0.05 & 0.431 & 254.47 & 0.067 \\ \hline
Synapse & U-Net & Seg-Grad-CAM & 0.1 & 0.348 & 203.74 & 0.067 \\ \hline
Synapse & U-Net & RISE & 0.2 & \textbf{0.935} & 319.11 & 52 \\ \hline
Synapse & U-Net & RISE & 0.4 & 0.656 & 120.76 & 52 \\ \hline
Synapse & U-Net & MSR & - & 0.797 & \textbf{11.43} & 3.38 \\ \hline
\hline
Synapse & TransUNet & Seg-Grad-CAM & 0.05 & 0.102 & 89.36 & 0.067 \\ \hline
Synapse & TransUNet & Seg-Grad-CAM & 0.1 & 0.082 & 74.86 & 0.067 \\ \hline
Synapse & TransUNet & RISE & 0.2 & \textbf{0.92} & 299.55 & 52 \\ \hline
Synapse & TransUNet & RISE & 0.4 & 0.595 & 118.08 & 52 \\ \hline
Synapse & TransUNet & MSR & - & 0.775 & \textbf{12.61} & 3.38 \\ \hline
\hline
COCO-2017 & DeepLabv3 & Seg-Grad-CAM & 0.05 & 0.693 & 5.47 & 0.067 \\ \hline
COCO-2017 & DeepLabv3 & Seg-Grad-CAM & 0.1 & 0.628 & 4.14 & 0.067 \\ \hline
COCO-2017 & DeepLabv3 & RISE & 0.2 & \textbf{0.883} & 9.2 & 52 \\ \hline
COCO-2017 & DeepLabv3 & RISE & 0.4 & 0.639 & 3.01 & 52 \\ \hline
COCO-2017 & DeepLabv3 & MSR & - & 0.81 & \textbf{0.22} & 3.38 \\ \hline
\end{tabular}%
}
\caption{Comparison between Minimally Sufficient Region's Approach vs. RISE vs. Seg-Grad-CAM on subsets of the Triangle dataset, Synaspe dataset, and COCO-2017 dataset. For MSR, the learning rate was kept at $0.1$, and $\lambda$ was kept at $0.01$. For RISE, $2000$ masks were used. For Seg-Grad-CAM, the layer of application was the bottleneck layer (the ASPP layer for the DeepLabv3 model). The total number of samples was 150, 400, and 181 for Triangle, Synapse, and COCO-2017 datasets respectively. For Dice explained, perturbation ratio, and the time, the mean value has been reported. The best results are reported in bold.}
\label{tab:comparisons}
\end{table*}

\subsection{Investigating the influence of the object size on the segmentation model}

\begin{figure*}
\begin{center}
    
    {\includegraphics[width=0.8\textwidth]{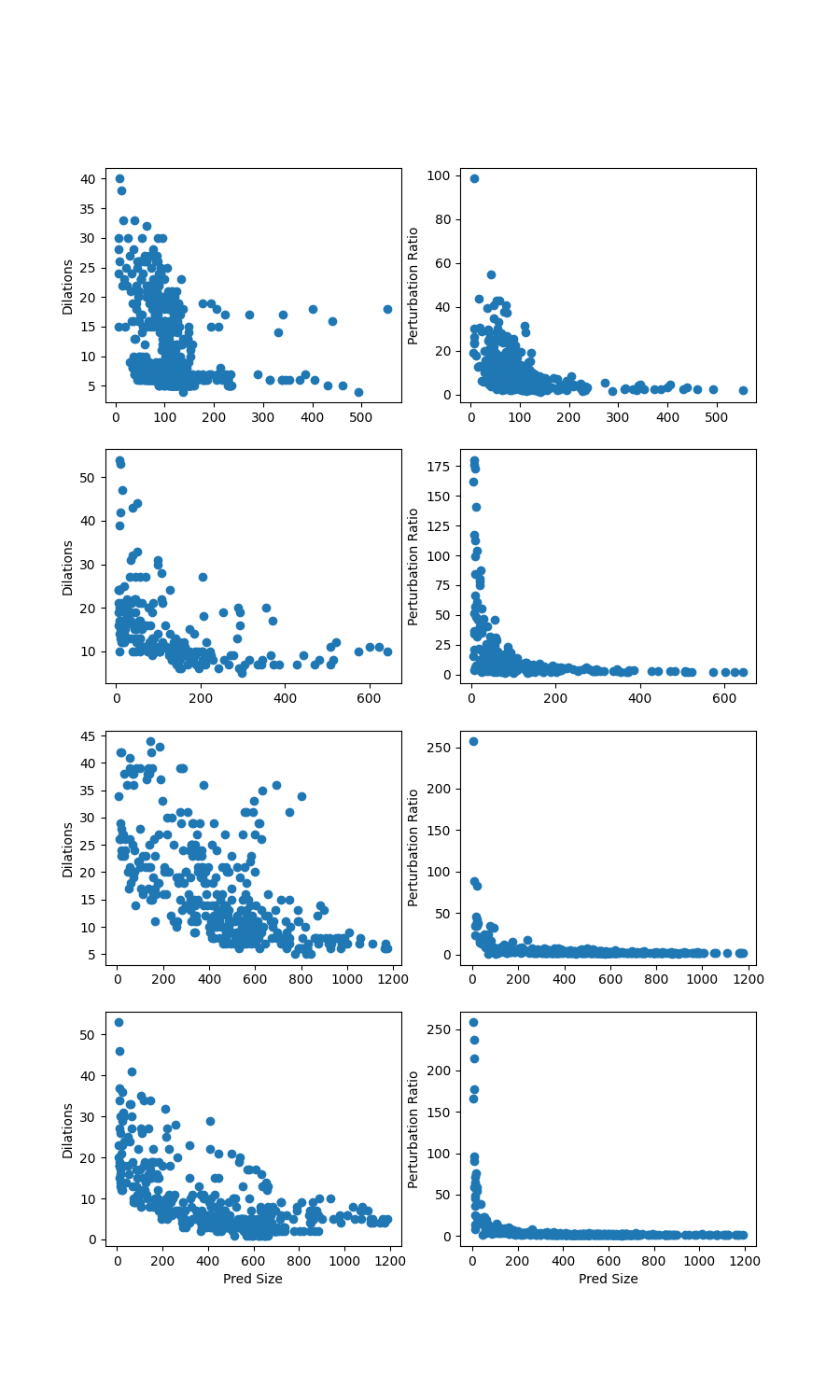}} \\

    \caption{Impact of Prediction Size on the Synapse multi-organ CT dataset for U-Net. Plots of the No. of Dilations against the Prediction size (left), and the Perturbation Ratio against the Prediction size (right). The plots, from top-to-bottom, are for the categories: Aorta, Gall Bladder, Left Kidney, Right Kidney.}
    
    \label{size_v_rest_unet}
    
\end{center}
\end{figure*}

\begin{figure*}
\begin{center}
    
    {\includegraphics[width=0.8\textwidth]{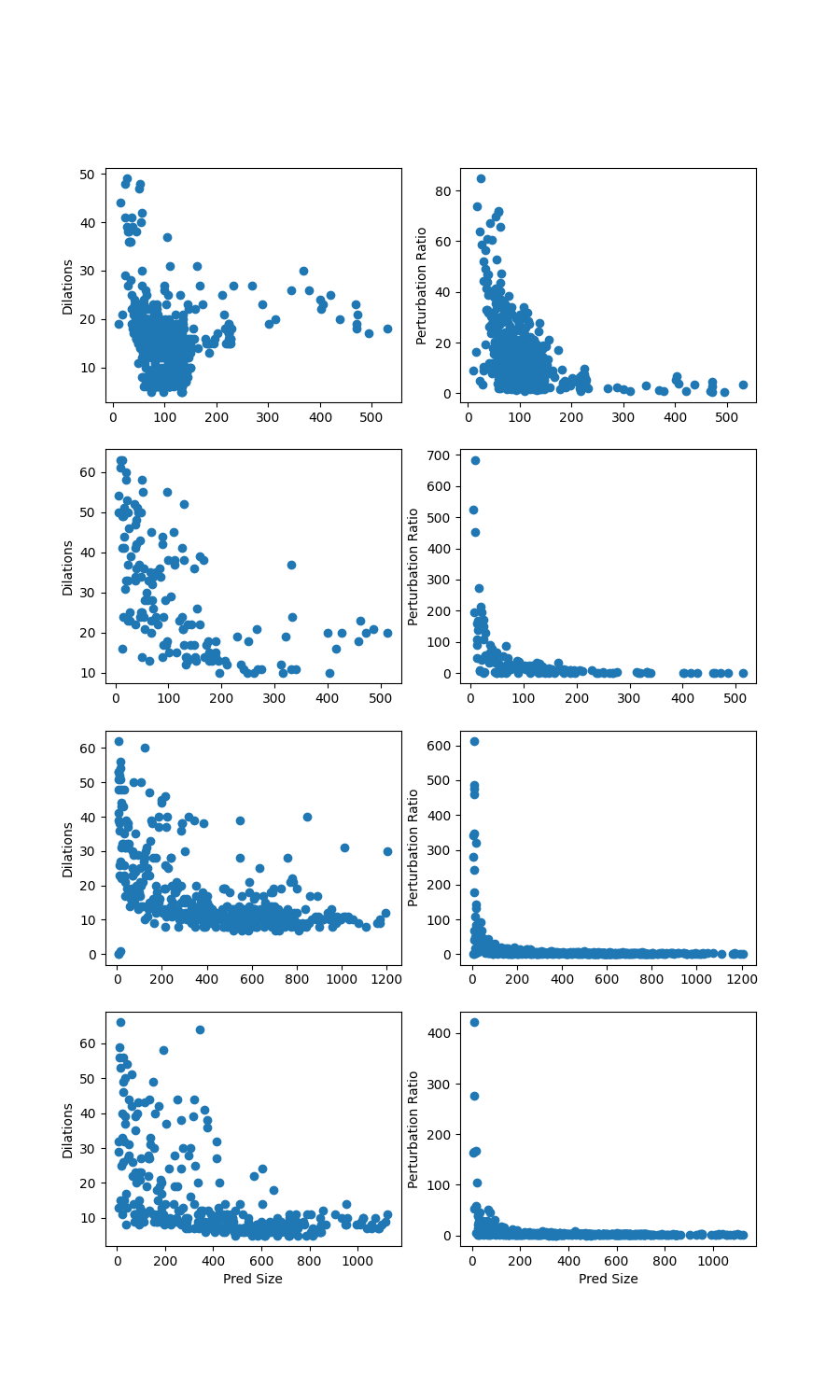}} \\

    \caption{Impact of Prediction Size on the Synapse multi-organ CT dataset for TransUNet. Plots of the No. of Dilations against the Prediction size (left), and the Perturbation Ratio against the Prediction size (right). The plots, from top-to-bottom, are for the categories: Aorta, Gall Bladder, Left Kidney, Right Kidney.}
    
    \label{size_v_rest_transunet}
    
\end{center}
\end{figure*}

\begin{figure*}
\begin{center}
    
    {\includegraphics[width=0.8\textwidth]{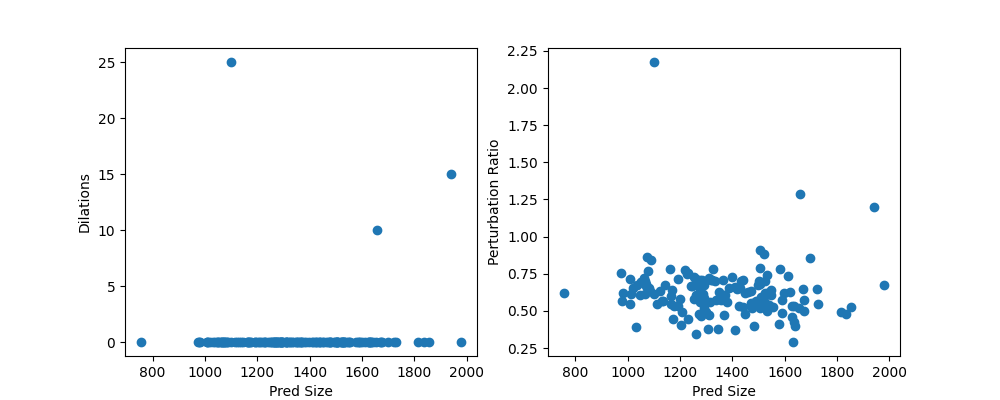}} \\
    \caption{Impact of Prediction Size on a subset of the Triangle dataset. Plot of the No. of Dilations against the Prediction size (left), and the Perturbation Ratio against the Prediction size (right).}
    
    \label{size_v_rest_florian}
    
\end{center}
\end{figure*}

\begin{figure*}
\begin{center}
    
    {\includegraphics[width=0.8\textwidth]{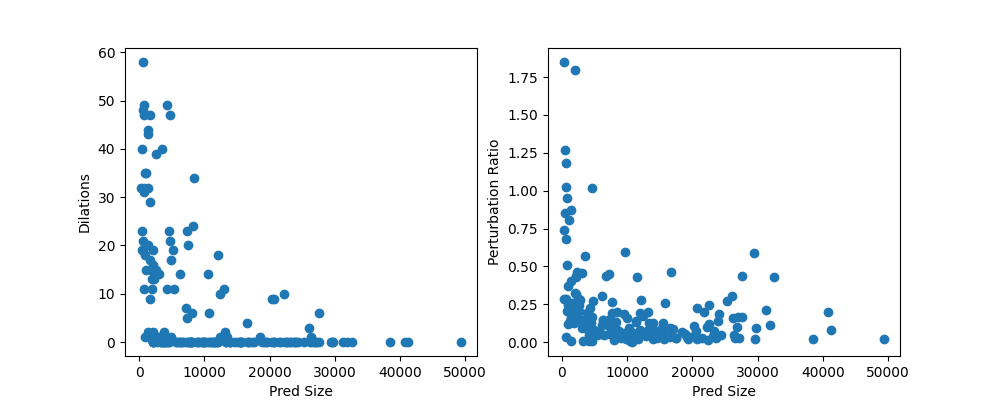}} \\
    \caption{Impact of Prediction Size on a subset of the COCO-2017 dataset. Plot of the No. of Dilations against the Prediction size (left), and the Perturbation Ratio against the Prediction size (right). Results from all eight categories are merged to achieve a single plot.}
    
    \label{size_v_rest_coco}
    
\end{center}
\end{figure*}

We further investigate the utility of saliency maps in terms of providing us with insights regarding the global process of segmentation, with respect to the size of the object to be segmented. We explore (i) the \textbf{number of dilations} required in order to arrive at the $X_{SR}$, and (ii) the \textbf{perturbation ratio}. Both of these are plotted against the prediction size. Figures. \ref{size_v_rest_unet} and \ref{size_v_rest_transunet} display these plots for the Synapse dataset for the U-Net and TransUNet respectively. Figures. \ref{size_v_rest_florian} and \ref{size_v_rest_coco} displays them for the Triangle dataset and COCO-2017 dataset respectively.

In general, we observe a decreasing trend for both the number of dilations as well as the perturbation ratio as the object's size increases for both the Synapse as well as COCO-2017 datasets. Additionally, we observe this trend for all three of our models: U-Net, DeepLabv3, and TransUNet. For a smaller prediction size, the perturbation ratio and the number of dilations are higher as opposed to a larger prediction size. This implies that the model seems to be requiring comparatively more visual data in order to segment smaller objects whereas as the size of the objects increase, the relative requirement of visual data decreases. In other words, the size of the object to be segmented and the relative amount of visual data required by the segmentation model to successfully segment it follows a roughly inverse relationship. 

The absence of this trend in the Triangle dataset can be attributed to its artificial nature. Given the similarity of objects to be segmented as opposed to the inherent diversity present in the medical and natural images datasets, the perturbation ratio as well as the number of dilations remained fairly constant across different input samples.

\subsection{Impact of Parameters}

We utilized a learning rate of 0.1 for the optimizer to find the MSR, a $\lambda$ of 0.01, and a mask size of 224 $\times$ 224 for our experiments. In the following two sections we report results for a variety of other parameter configurations in order to justify our choice.

\subsubsection{Learning rate vs. $\lambda$}

We report results for 9 pairs of learning rate and $\lambda$ such that each parameter takes on a value of 0.001, 0.01, and 0.1. Results for the Triangle dataset are displayed in Table. \ref{lr_l1_florian}, results for the Synapse dataset are displayed in Tables \ref{lr_l1_synapse_unet} and \ref{lr_l1_synapse_transunet} for U-Net and TransUNet respectively, and results for the COCO-2017 dataset are displayed in Table \ref{lr_l1_coco}. It is immediately obvious that learning rate and $\lambda$ are significant parameters for our saliency generation method. The best results are attributable to a learning rate of 0.1 (rightmost column) in each table. For this learning rate, $\lambda$ of 0.1 returns the best results for \textbf{perturbation ratio}, but it seems that it does so at the expense of \textbf{Dice explained}. Between 0.001 and 0.01, a $\lambda$ of 0.01 consistently provides the best balance between our two metrics. 

\begin{table*}[t]
\begin{tabularx}{\textwidth}{l@{\hskip 1in}c@{\hskip 0.5in}c}
    \begin{tabular}{ccccc}
         \multicolumn{5}{c}{\textbf{Dice Explained}} \\   
         &  &  &  & \\
         & & \multicolumn{3}{c}{\textbf{learning rate}} \\
         \multirow{4}{*}{\textbf{$\lambda$}} & & & \\
         & &  0.001 & 0.01 & 0.1 \\ \cline{3-5}
         & 0.001 & \textbf{0.987} & 0.977 & 0.969 \\ \cline{3-5}
         & 0.01  & \textbf{0.987} & 0.976 & 0.965 \\ \cline{3-5}
         & 0.1  & \textbf{0.987} & 0.976 & 0.924 \\ \cline{3-5}
         
    \end{tabular} &

    \begin{tabular}{ccccc}
        \multicolumn{5}{c}{\textbf{Perturbation Ratio}} \\   
         &  &  &  & \\
         & & \multicolumn{3}{c}{\textbf{learning rate}} \\
         \multirow{4}{*}{\textbf{$\lambda$}} & & & \\
         & &  0.001 & 0.01 & 0.1 \\ \cline{3-5}
         & 0.001 & 0.961 & 0.96 & 0.927 \\ \cline{3-5}
         & 0.01  & 0.961 & 0.96 & 0.627 \\ \cline{3-5}
         & 0.1  & 0.961 & 0.957 & \textbf{0.232} \\ \cline{3-5}
    \end{tabular} \\

        \\
        
\end{tabularx}
\caption{Impact of learning rate and $\lambda$ on a subset of the Triangle Dataset. The mask size was kept at 224 $\times$ 224. The total number of samples was 150. The left table reports the average Dice explained whereas the right table reports the average perturbation ratio. The best results are reported in bold.}
\label{lr_l1_florian}
\end{table*}

\begin{table*}
\begin{tabularx}{\textwidth}{l@{\hskip 1in}c@{\hskip 0.5in}c}
    \begin{tabular}{ccccc}
         \multicolumn{5}{c}{\textbf{Dice Explained}} \\   
         &  &  &  & \\
         & & \multicolumn{3}{c}{\textbf{learning rate}} \\
         \multirow{4}{*}{\textbf{$\lambda$}} & & & \\
         & &  0.001 & 0.01 & 0.1 \\ \cline{3-5}
         & 0.001 & 0.771 & 0.773 & 0.794 \\ \cline{3-5}
         & 0.01  & 0.768 & 0.774 & \textbf{0.797} \\ \cline{3-5}
         & 0.1  & 0.749 & 0.78 & 0.74 \\ \cline{3-5}
         
    \end{tabular} &

    \begin{tabular}{ccccc}
        \multicolumn{5}{c}{\textbf{Perturbation Ratio}} \\   
         &  &  &  & \\
         & & \multicolumn{3}{c}{\textbf{learning rate}} \\
         \multirow{4}{*}{\textbf{$\lambda$}} & & & \\
         & &  0.001 & 0.01 & 0.1 \\ \cline{3-5}
         & 0.001 & 140.123 & 109.974 & 14.045 \\ \cline{3-5}
         & 0.01  & 140.123 & 97.405 & 11.431 \\ \cline{3-5}
         & 0.1  & 140.123 & 53.767 & \textbf{5.403} \\ \cline{3-5}
    \end{tabular} \\

        \\
        
\end{tabularx}

\caption{Comparison between impact of learning rate and $\lambda$ on a subset of the Synapse multi-organ CT Dataset. The mask size was kept at 224 $\times$ 224. The total number of samples was 400 (50 samples per class). The left table reports the average Dice explained whereas the right table reports the average perturbation ratio. The best results are reported in bold.}

\label{lr_l1_synapse_unet}
\end{table*}

\begin{table*}
\begin{tabularx}{\textwidth}{l@{\hskip 1in}c@{\hskip 0.5in}c}
    \begin{tabular}{ccccc}
         \multicolumn{5}{c}{\textbf{Dice Explained}} \\   
         &  &  &  & \\
         & & \multicolumn{3}{c}{\textbf{learning rate}} \\
         \multirow{4}{*}{\textbf{$\lambda$}} & & & \\
         & &  0.001 & 0.01 & 0.1 \\ \cline{3-5}
         & 0.001 & 0.79 & \textbf{0.796} & 0.767 \\ \cline{3-5}
         & 0.01  & 0.788 & \textbf{0.796} & 0.75 \\ \cline{3-5}
         & 0.1  & 0.773 & 0.783 & 0.68 \\ \cline{3-5}
         
    \end{tabular} &

    \begin{tabular}{ccccc}
        \multicolumn{5}{c}{\textbf{Perturbation Ratio}} \\   
         &  &  &  & \\
         & & \multicolumn{3}{c}{\textbf{learning rate}} \\
         \multirow{4}{*}{\textbf{$\lambda$}} & & & \\
         & &  0.001 & 0.01 & 0.1 \\ \cline{3-5}
         & 0.001 & 158.849 & 117.877 & 14.861 \\ \cline{3-5}
         & 0.01  & 158.849 & 104.78 & 12.742 \\ \cline{3-5}
         & 0.1  & 158.849 & 56.32 & \textbf{4.571} \\ \cline{3-5}
    \end{tabular} \\

        \\
        
\end{tabularx}

\caption{Comparison between impact of learning rate and $\lambda$ on a subset of the Synapse multi-organ CT Dataset. The mask size was kept at 224 $\times$ 224. The total number of samples was 400 (50 samples per class). The left table reports the average Dice explained whereas the right table reports the average perturbation ratio. The best results are reported in bold.}

\label{lr_l1_synapse_transunet}
\end{table*}

\begin{table*}
\begin{tabularx}{\textwidth}{l@{\hskip 1in}c@{\hskip 0.5in}c}
    \begin{tabular}{ccccc}
         \multicolumn{5}{c}{\textbf{Dice Explained}} \\   
         &  &  &  & \\
         & & \multicolumn{3}{c}{\textbf{learning rate}} \\
         \multirow{4}{*}{\textbf{$\lambda$}} & & & \\
         & &  0.001 & 0.01 & 0.1 \\ \cline{3-5}
         & 0.001 & 0.583 & 0.737 & \textbf{0.824} \\ \cline{3-5}
         & 0.01  & 0.583 & 0.739 & 0.81 \\ \cline{3-5}
         & 0.1  & 0.575 & 0.736 & 0.751 \\ \cline{3-5}
         
    \end{tabular} &

    \begin{tabular}{ccccc}
        \multicolumn{5}{c}{\textbf{Perturbation Ratio}} \\   
         &  &  &  & \\
         & & \multicolumn{3}{c}{\textbf{learning rate}} \\
         \multirow{4}{*}{\textbf{$\lambda$}} & & & \\
         & &  0.001 & 0.01 & 0.1 \\ \cline{3-5}
         & 0.001 & 3.478 & 2,749 & 0.33 \\ \cline{3-5}
         & 0.01  & 3.478 & 2.57 & 0.215 \\ \cline{3-5}
         & 0.1  & 3.478 & 1.065 & \textbf{0.061} \\ \cline{3-5}
    \end{tabular} \\

        \\
        
\end{tabularx}

\caption{Comparison between impact of learning rate and $\lambda$ on a subset of the COCO-2017 Dataset. The mask size was kept at 224 $\times$ 224. The total number of samples was 181. The left table reports the average Dice explained whereas the right table reports the average perturbation ratio. The best results are reported in bold.}

\label{lr_l1_coco}
\end{table*}

\subsubsection{Impact of Mask Size}

We report results for 3 mask sizes ranging from 56 $\times$ 56, 112 $\times$ 112, and 224 $\times$ 224. Results for the Triangle, Synapse, and COCO-2017 datasets are displayed in Table. \ref{tab:mask-size}. Once again it is clear that mask size is an important parameter, more so in the domain of \textbf{perturbation ratio} as compared to \textbf{Dice explained}. Overall, our best results are obtained with a mask size of 224 $\times$ 224. 


\begin{table*}[]
\centering
\resizebox{1\columnwidth}{!}{%
\begin{tabular}{|c|c|c|c|c|}
\hline
\textbf{Dataset} & \textbf{Model} & \textbf{Mask Size} & \textbf{Dice Explained} & \textbf{Perturbation Ratio} \\ \hline
Triangle & U-Net & 56 & \textbf{0.975} & 2.212 \\ \hline
Triangle & U-Net &  112 & 0.971 & 1.56 \\ \hline
Triangle & U-Net &  224 & 0.965 & \textbf{0.627} \\ \hline
\hline
Synapse & U-Net &  56 & 0.769 & 68.078 \\ \hline
Synapse & U-Net &  112 & 0.783 & 49.478 \\ \hline
Synapse & U-Net &  224 & \textbf{0.797} & \textbf{11.431} \\ \hline
\hline
Synapse & TransUNet &  56 & 0.674 & 67.807 \\ \hline
Synapse & TransUNet & 112 & 0.729 & 52.442 \\ \hline
Synapse & TransUNet & 224 & \textbf{0.75} & \textbf{12.742} \\ \hline
\hline
COCO-2017 & DeepLabv3 & 56 & 0.77 & 1.97 \\ \hline
COCO-2017 & DeepLabv3 & 112 & 0.808 & 1.576 \\ \hline
COCO-2017 & DeepLabv3 & 224 & \textbf{0.81} & \textbf{0.215} \\ \hline
\end{tabular}%
}
\caption{Comparison between impact of mask sizes on subsets of the Triangle dataset, Synapse dataset, and COCO-2017 dataset. The learning rate was kept at 0.1, and the $\lambda$ was kept at 0.01. The total number of samples was 150, 400, and 181 for Triangle, Synapse, and COCO-2017 datasets respectively. For both the Dice explained as well as the perturbation ratio, the mean value has been reported. The best results are reported in bold}
\label{tab:mask-size}
\end{table*}

\subsection{Post-Hoc Assessment of Segmentation Model Reliability}

\begin{table*}[]
\centering
\resizebox{0.8\columnwidth}{!}{%
\begin{tabular}{|c|c|c|c|c|}
\hline
\textbf{Class} & \textbf{No. of Samples} & \textbf{Post-hoc Accuracy} & \textbf{Post-hoc AUC} \\ \hline
Aorta & 540 & 0.78 & 0.78 \\ \hline
Gall Bladder & 147 & 0.91 & 0.95 \\ \hline
Left Kidney & 242 & 0.95 & 0.98 \\ \hline
Right Kidney & 260 & 0.93 & 0.97 \\ \hline
Liver & 422 & 0.93 & 0.94 \\ \hline
Pancreas & 217 & 0.98 & 0.72 \\ \hline
Spleen & 255 & 0.78 & 0.88 \\ \hline
Stomach & 292 & 0.77 & 0.88 \\ \hline
\end{tabular}%
}
\caption{Post-hoc Model reliability results on eight classes from the Synapse multi-organ dataset}
\label{tab:post-hoc}
\end{table*}

\begin{figure*}
\centering
    \includegraphics[width=0.5\textwidth]{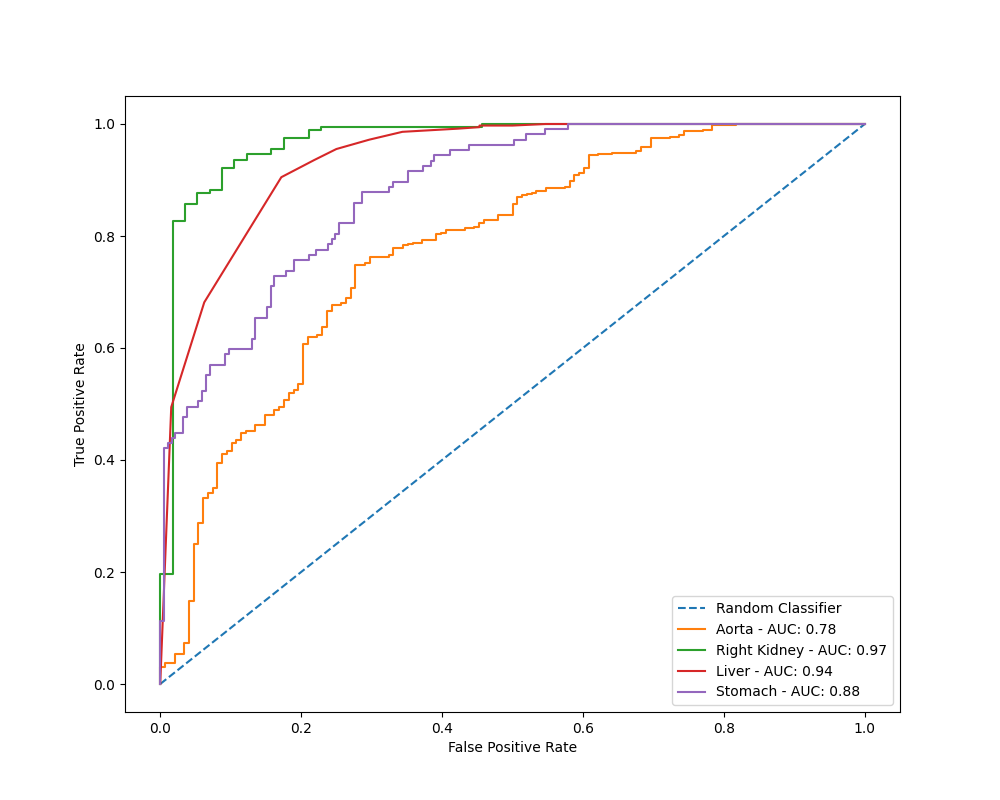} 
    \caption{ROC curves of post-hoc reliability classifiers trained for the 'Aorta', 'Right Kidney', 'Liver', and 'Stomach' categories respectively.}
    \label{auc}
\end{figure*}

\begin{figure*}
\centering
    \includegraphics[width=1\textwidth]{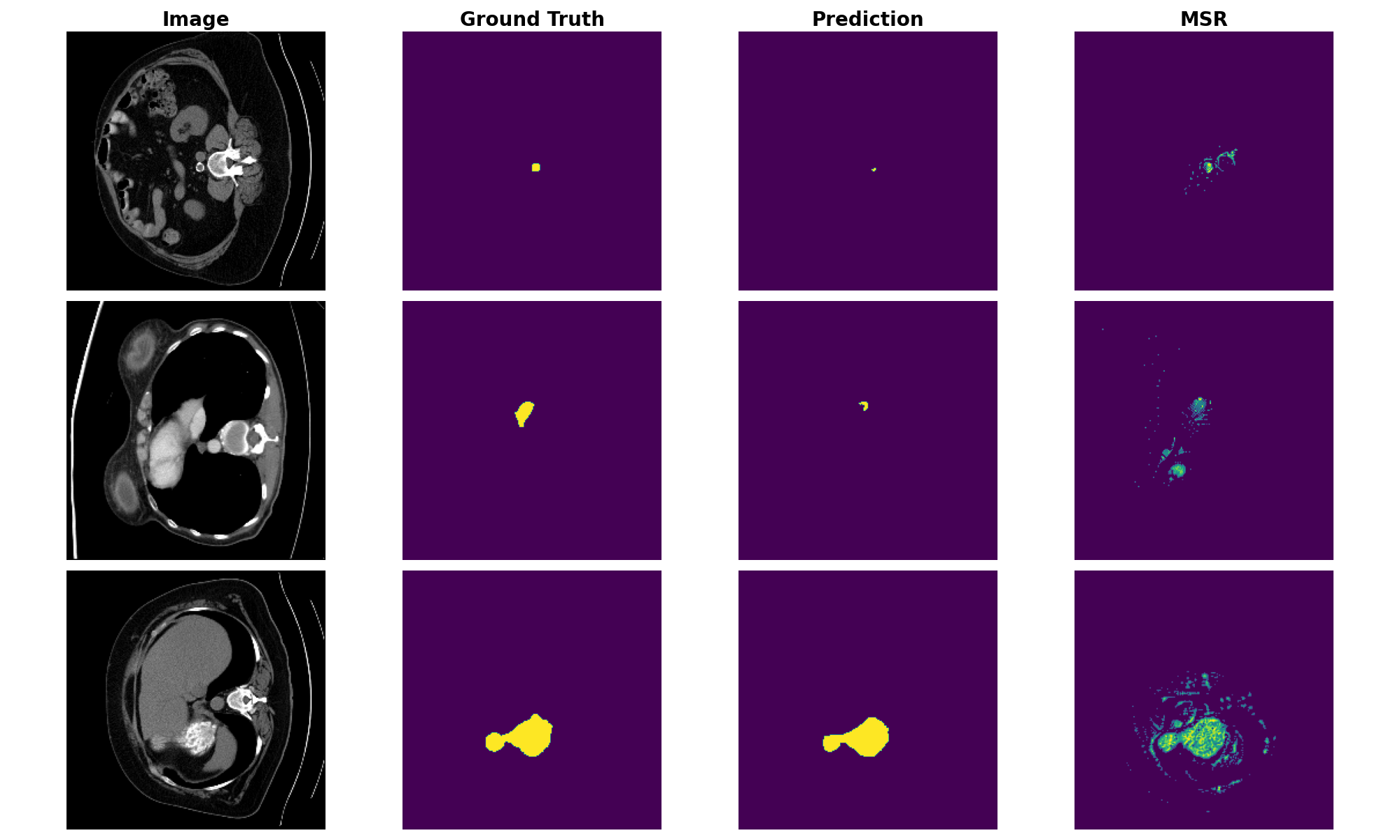} 
    \caption{Application of the post-hoc reliability classifier (a simple logistic regression model) on three examples from the Synapse dataset. From top to bottom, the categories are 'Aorta', 'Liver', and 'Stomach'. The post-hoc reliability classifier is correctly able to identify incorrect predictions from the segmentation model for the first and second row as well as the correct prediction from the segmentation model in the third row.}
    \label{post-hoc-img}
\end{figure*}

Where saliency maps are a useful and informative tool for the end user, a potentially useful line of study is to identify whether these saliency maps can, in one way or another, be related to the segmentation model's actual performance. A possible example of this can be a case where saliency maps generated for incorrect model predictions and those generated for correct model predictions follow different patterns. If this ends up being the case, a discriminative classifier can be trained using saliency maps as features which can act as a proxy for post-hoc model reliability. For any new image, the model's prediction as well as the saliency map generated for that prediction can be fed into the discriminative classifier from which one could get a certain confidence as to whether this prediction can be considered correct or not. 

We utilize three features from our saliency generation process namely (i) the number of dilations required to generate $X_{SR}$, (ii) the Dice score between the model's prediction on the original image and the model's prediction on the minimially sufficient region, and (iii) the ratio of the number of non-zero pixels in the minimally sufficient region to the the number of non-zero pixels in the model's prediction of the object of interest. Given that we already have access to the ground truth Dice values, we train a logistic regression classifier to predict whether the model's prediction as compared with the ground truth would cross a Dice threshold of $0.9$. Results for individual categories can be seen in Table. \ref{tab:post-hoc} whereas the ROC curves for a subset of these categories can be seen in Figure. \ref{auc}.

Unlike image classification where the prediction is binary (either it belongs to the category or it does not), image segmentation is a task involving dense prediction where a meaningful overlap between the ground truth and the segmentation model's prediction is entirely subjective. It is possible that an overlap of $0.5$ is useful enough in certain medical settings, and an overlap of anything less than $0.9$ unworthy of consideration in high precision industrialized settings. If saliency maps show the potential to offer assistance in such situations as proxy indicators of the model's performance in the real world, they can act as an additionally informative feature for the end users, and for certain cases might even offer the potential of automating the pipeline with regards to which of the model's predictions to accept and which one's to reject.

Figure \ref{post-hoc-img} shows three examples to demonstrate the efficacy of our technique. In the first row, the Dice between the aorta as predicted by the segmentation model and the ground truth is clearly less than $0.9$. Our post-hoc classifier correctly identifies this with a probability 0f \textbf{$0.71$}. In the second row, once again, the Dice between the liver as predicted by the segmentation model and the ground truth is less than $0.9$ and our post-hoc classifier is able to identify this with a probability of \textbf{$0.72$}. In the third row, the Dice between the stomach as predicted by the segmentation model and the ground truth is above $0.9$, and our post-hoc classifier is able to identify this with a probability of \textbf{$0.92$}. Given the absence of ground truth during test time, such a post-hoc classifier can increase the practitioner's trust on the segmentation model's predictions.

\subsection{Comparison}

The proposed method offers multiple advantages over other existing methods. If we compare it with Seg-Grad-CAM, the minimally sufficient region's approach is model-agnostic whereas Grad-CAM does not fare well for transformer based models due to its explicit reliance on receptive field. Additionally, for Seg-Grad-CAM, the choice of the model's layer is not straightforward, and even though most users apply it to the bottleneck layer of the segmentation model, it is not immediately clear as to why the other layers of the decoder are being ignored. RISE, on the other hand, is a model-agnostic method as well, but it comes with a cumbersome computational cost due to the number of masks it requires to be generated which leads to it working on the order of minutes on a single image whereas the minimally sufficient region's method works on the order of seconds. Also, RISE gives us coarse explanations whereas the proposed method gives us both coarse grained (sufficient region) and fine grained explanations (minimally sufficient regions). 

\subsection{Limitations}

As the method is based on optimization, it naturally involves determining a number of hyperparameters such as $\lambda$, $\alpha_{0}$, $\alpha_{l}$, learning rate, mask size, and the number of iterations. The most important of these are $\lambda$ and learning rate as high values for both threaten to destroy the mask entirely whereas low values would make it appear as if no pruning has been done to the mask at all. Additionally, mask size is another important hyperparameter with bigger masks ending up being more pruned as compared to their smaller counterparts. Multiple experiments had to be performed in order to determine these hyperparameters. Also, as the algorithm is based on a perturbation based strategy, it is iterative in nature making it relatively slow as compared to some of the other explainability algorithms such as Seg-Grad-CAM which only require a single forward and backward pass.

\section{Conclusion}

This work proposes a simple, two stage, model agnostic method,  MiSuRe (\textbf{Mi}nimally \textbf{Su}fficient \textbf{Re}gion), to generate saliency maps for image segmentation. The first step, motivated by the inherent bias in image segmentation, is to dilate a mask focusing on the object of interest in order to identify a sufficient region $X_{SR}$. This sufficient region then serves as the initialization of an optimization process in which the goal is to further prune it in order to arrive at the minimally sufficient region ($X_{MSR}$). While the sufficient region can be considered as a coarser explanation, the minimally sufficient region provides us with a finer one. This approach can be utilized to extract insights from the segmentation process as a whole, an example of which is to plot the relative size of the visual content required by the segmentation model against the size of the object to be segmented. For the examples considered, this plot reveals an inverse relationship between the two. Additionally, a potential application for this approach is to utilize features obtained from saliency maps in order to train discriminative classifiers which would act as a proxy for the model's predictions on unseen future data. Compared to some of the existing methods of generating saliency maps for image segmentation, the current approach stands-out as being model agnostic, computationally feasible, and providing the end user with a coarse as well as a fine explanation.


\bibliographystyle{unsrtnat}
\bibliography{references}  






\end{document}